\documentclass[letterpaper, 10 pt, conference]{ieeeconf}
\IEEEoverridecommandlockouts
\usepackage{cite}
\usepackage{amsmath,amsfonts,amssymb}
\usepackage{algorithmic}
\usepackage{array}
\usepackage{textcomp}
\usepackage{stfloats}
\usepackage{url}
\usepackage{color}
\usepackage{float}
\usepackage{verbatim}
\usepackage{graphicx}
\usepackage{multirow}
\usepackage{gensymb}
\usepackage{makecell}
\usepackage{subfigure}
\usepackage{booktabs}
\usepackage{epstopdf}

\title{CM-Bench: A Comprehensive Cross-Modal Feature Matching Benchmark Bridging Visible and Infrared Images}

\author{Liangzheng Sun$^{1}$,
        Mengfan He$^{2}$,
        Xingyu Shao$^{2}$,
        Binbin Li$^{1}$,
        Zhiqiang Yan$^{1}$,
        Chunyu Li$^{2}$,
        Ziyang Meng$^{2}$,
        Fei Xing$^{2}$%
\thanks{Liangzheng Sun and Mengfan He contributed equally to this work. Corresponding author: Ziyang Meng.}%
\thanks{$^{1}$School of Instrument Science and Opto-Electronics Engineering, Beijing Information Science and Technology University, Beijing 100192, China
        {\tt\small 2023030031@bistu.edu.cn; 2024020382@bistu.edu.cn;\newline
        2024020267@bistu.edu.cn}}%
\thanks{$^{2}$Department of Precision Instrument, Tsinghua University, Beijing 100084, China
        {\tt\small hmf21@mails.tsinghua.edu.cn; shao-xy21@mails.tsinghua.edu.cn;\newline
        lcyfly1@163.com;\newline
        ziyangmeng@mail.tsinghua.edu.cn;\newline
        xingfei@mail.tsinghua.edu.cn}}%
}

\begin{document}
\maketitle
\thispagestyle{empty}
\pagestyle{empty}
\begin{abstract}

Infrared-visible (IR-VIS) feature matching plays an essential role in cross-modality visual localization, navigation and perception. 
Along with the rapid development of deep learning techniques, a number of representative image matching methods have been proposed. 
However, cross-modal feature matching is still a challenging task due to the significant appearance difference.
A significant gap for cross-modal feature matching research lies in the absence of standardized benchmarks and metrics for evaluations.
In this paper,  we introduce a comprehensive cross-modal feature matching benchmark, CM-Bench, which encompasses 30 feature matching algorithms across diverse cross-modal datasets.
Specifically, state-of-the-art traditional and deep learning-based methods are first summarized and categorized into sparse, semi-dense, and dense methods.
These methods are evaluated by different tasks including homography estimation, relative pose estimation, and feature-matching-based geo-localization.
In addition, we introduce a classification-network-based adaptive preprocessing front-end that automatically selects suitable enhancement strategies before matching.
We also present a novel infrared-satellite cross-modal dataset with manually annotated ground-truth correspondences for practical geo-localization evaluation.
The dataset and resource will be available at: https://github.com/SLZ98/CM-Bench.

\end{abstract}

\section{Introduction}
Infrared (IR) and visible (VIS) image sensors are widely used in numerous realistic applications, such as security monitoring \cite{qin2019infrared}, remote sensing \cite{zhao2024visible} and autonomous vehicles \cite{jiang2022thermal}.
They provide different but complementary modal information.
and therefore leveraging cross-modal information is essential for achieving a robust system.
In particular, the integration of cross-modal sensing information requires reliable data association through feature matching.
while existing feature matching methods primarily focus on algorithm design tailored for visible image sensors.
It remains unclear whether these methods are still feasible for cross-modal feature matching task.
Specifically, traditional feature matching algorithms are prone to fail because of their reliance on intensity or texture similarity.
On the other hand, deep-learning methods have demonstrated strong capabilities in learning discriminative features.
However, their generalization to cross-modal tasks remains largely under-explored.
Additionally, algorithms specially designed for cross-modal matching are often evaluated in a constrained dataset and lack of fully comparison.
To the best of our knowledge, the systematic evaluation for cross-modal feature matching has not been explored, from a comprehensive collection of datasets to consistent evaluation metrics.

\begin{figure}[t]
    \centering
    \includegraphics[width=1\linewidth]{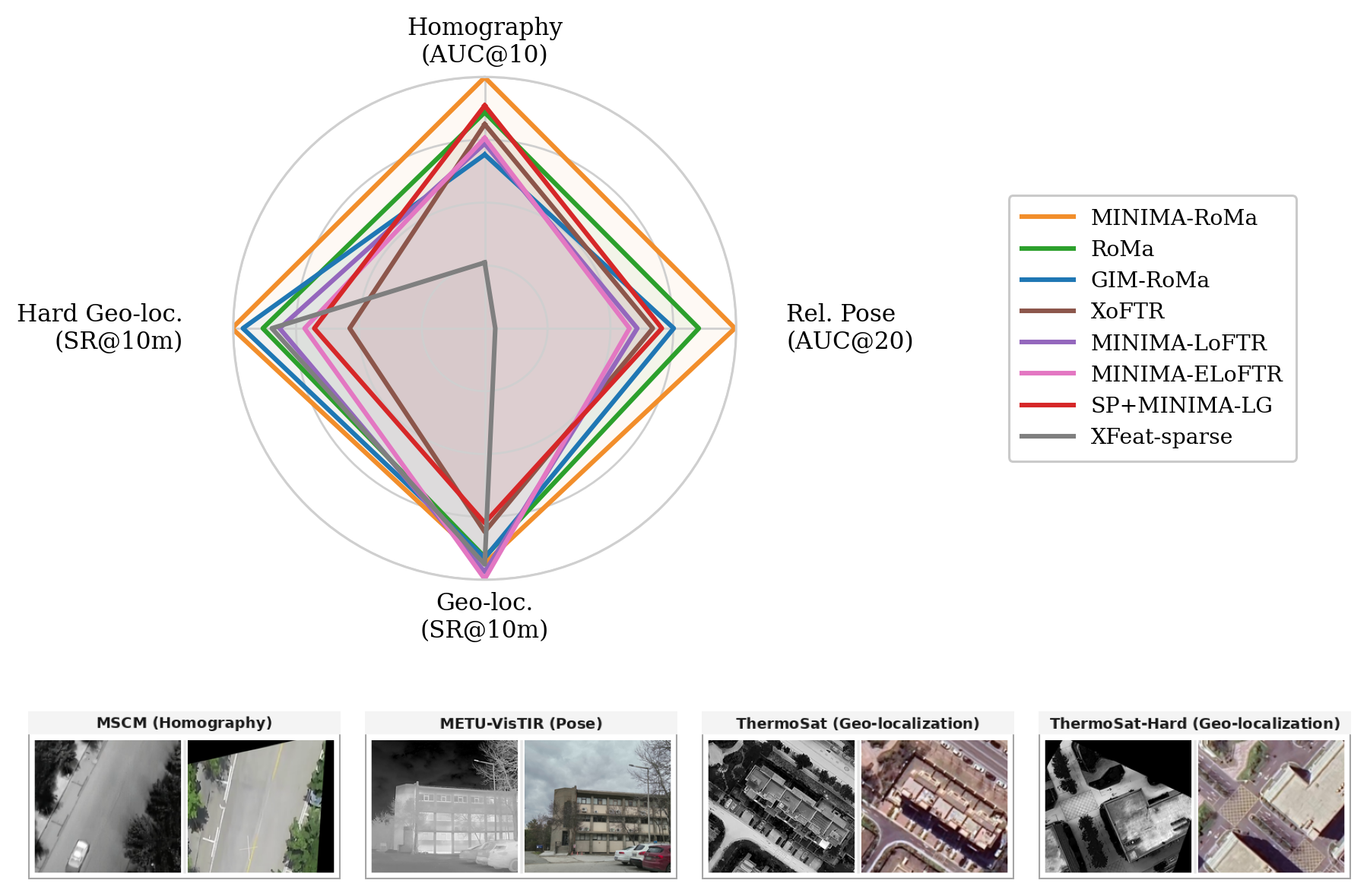}
    \vspace{-3.2ex}
    \caption{Overview of eight representative matchers across four benchmark tasks, together with representative samples from the corresponding evaluation settings.}
    \vspace{-3.6ex}
    \label{fig:figure1_overall}
\end{figure}

Another major challenge in advancing IR-VIS feature matching lies in the limited availability of high-quality datasets tailored for cross-modal tasks.
Existing cross-modal datasets are often either too small or lack diversity in environmental settings and viewpoints to support effective evaluation.
Furthermore, in addition to cross-modal feature matching technique, the image preprocessing are also promising engineering solutions to improve cross-modal feature matching performance, including contrast enhancement and edge feature extraction.
% Addressing these challenges also requires a comprehensive benchmark for consistent evaluation.

In this work, we present the first comprehensive cross-modal feature matching benchmark, named \textbf{CM-Bench}, which contains 30 feature matching algorithm evaluations in total, including traditional handcrafted features (e.g., SIFT\cite{sift}, RIFT \cite{rift}), advanced deep learning-based methods (e.g. SuperPoint \cite{superpoint}, ALIKED \cite{aliked}), and methods specifically designed for cross-modal matching (e.g. XoFTR \cite{tuzcuouglu2024xoftr}, MINIMA \cite{jiang2024minima}).
Unlike the existing benchmarks designed for the RGB-only feature matching works \cite{yu2024gv, jin2021image, bonilla2024mismatched}, CM-Bench presents a clarified evaluation for the cross-modal matching task.
The proposed benchmark is based on utilizing a large number of publicly available datasets and self-collected data for comprehensive evaluation. Fig.~\ref{fig:figure1_overall} provides an overview of performance comparisons among state-of-the-art methods across different evaluation tasks in our benchmark.
The contributions are summarized as follows:

\begin{itemize}
\item We propose \textbf{CM-Bench}, a comprehensive comparison of \textbf{30} feature matching algorithms across \textbf{10} diverse cross-modal datasets.
The evaluated algorithms include various types of methods, spanning from classic handcrafted features to the state-of-the-art deep learning models.
The selected datasets capture a variety of environments, sensor settings, and collection conditions to reflect realistic challenges.

\item We present quantitative and qualitative evaluations on algorithm performance, utilizing several effective metrics.
Furthermore, we introduce a classification-network-based adaptive preprocessing front-end, which automatically selects suitable enhancement strategies for different image pairs before matching.

\item We release a new  cross-modal dataset between infrared and satellite images, termed the Aerial Thermal--Satellite Localization Dataset (ThermoSat).
The infrared images and corresponding satellite images are manually calibrated and annotated with ground-truth points. The current benchmark uses 832 annotated pairs captured by a DJI Matrice 4T platform, covering representative agricultural and urban scenes.

\end{itemize}

%%%%%%%%%%%%%%%%%%%%%%%%%%%%%%%%%%%%%%%%%%%%%%%%%%%%%%%%%%%%%%%%%%%%%%%%%%%%%%%%

\label{sec:related_work}
\section{Related Work}

\subsection{RGB-Only Image Feature Matching}
Feature matching is a fundamental problem in computer vision, and many methods have been proposed for RGB-only images.
These methods can be categorized into three classes based on the matching granularity: sparse keypoint-based matching, semi-dense correspondence methods, and dense correspondence techniques, which respectively focus on distinctive keypoints, larger image regions, and pixel-level correspondence.

\subsubsection{Sparse Keypoint-Based Matching}
% Sparse keypointbased methods extract significant local points in images and establish correspondences using descriptors.
Traditional handcrafted methods, such as SIFT \cite{sift}, ORB \cite{orb}, and SURF \cite{surf}, are developed for single-modal feature matching, but they often struggle in IR-VIS tasks because the assumption of consistent intensity patterns does not hold across modalities. Learning-based features such as SuperPoint \cite{superpoint}, ALIKED \cite{aliked}, and R2D2 \cite{r2d2}, together with refinement techniques such as SuperGlue \cite{sarlin2020superglue}, LightGlue \cite{lindenberger2023lightglue}, and Patch2Pix \cite{patch2pix}, substantially improve single-modality matching performance.

\subsubsection{Semi-Dense Correspondence}
% Semi-dense methods expand the matching scope beyond sparse keypoints to the broader image regions. 
One of the most popular approaches of semi-dense methods is LoFTR\cite{loftr}, which directly predicts dense correspondences between pixels without keypoint detection.
Following this idea, Efficient LoFTR \cite{wang2024eloftr} optimizes the previous pipeline for optimal computational efficiency without performance loss.
Other methods, including ASpanFormer\cite{chen2022aspanformer} and MatchFormer\cite{matchformer}, extend transformer-based architectures to refine matching through hierarchical attention mechanisms and cross-scale learning.
% All of these methods rely on transformer-based architectures, which can model global and local dependencies for effective feature alignment.

\subsubsection{Dense Correspondence}
% The dense correspondence method provides pixel-wise matching across the entire image, aiming for the most comprehensive coverage among all the approaches.
% This kind of methods align the entire images based on global feature properties, instead of relying on the selected points.
DKM \cite{edstedt2023dkm} and GIM-DKM \cite{gim} are two representative dense correspondence solutions that capture detailed global relationships and align entire images based on global feature properties.
% but require significant computational resources due to their pixel-wise processing.
Other dense matching algorithms, such as RoMa\cite{edstedt2024roma} and RoMa v2\cite{edstedt2025roma}, focus on geometric consistency to provide robust dense correspondence across different scales. 
On the other hand, DUSt3R\cite{wang2024dust3r} and MASt3R\cite{mast3r} integrate spatial and structural constraints by directly predicting geometric relationships under viewpoint changes, although dense matching methods are often computationally demanding.

\subsection{Cross-Modal Feature Matching}
Algorithms for RGB cameras may not excel in cross-modal matching due to generalization limits, prompting the development of task-specific cross-modal matching algorithms.
A representative handcrafted cross-modal feature is RIFT \cite{rift}, which utilizes multi-scale gradient orientations to ensure invariance to radiation changes.
XoFTR\cite{tuzcuouglu2024xoftr} is a prominent semi-dense algorithm designed specifically for addressing modality shifts by generating pseudo infrared images during training, while MINIMA~\cite{jiang2024minima} uses generative models to synthesize infrared data for fine-tuning.

%%%%%%%%%%%%%%%%%%%%%%%%%%%%%%%%%%%%%%%%%%%%%%%%%%%%%%%%%%%%%%%%%%%%%%%%%%%%%%%%
\section{Methodology}
\label{sec:methodology}
\subsection{Cross-Modal Feature Matching and Datasets}
The modalities between infrared and visible exhibit substantial appearance differences, including variations in texture, contrast, edge responses and spectrum.
These characteristics often cause traditional feature matching algorithms designed for single-modality scenarios to perform poorly.
This study establishes a comprehensive evaluation framework to systematically assess the performance of various feature matching algorithms in IR-VIS cross-modal scenarios.

Our evaluation framework utilizes three benchmark groups. First, for homography estimation, we use the Multi-Source Cross-Modal dataset (MSCM), containing 7,083 pixel-aligned infrared-visible image pairs from eight existing datasets: (1) VisDrone, providing diverse aerial urban scenes captured from drone platforms; (2) DUT-VTUAV~\cite{zhang2022visible}, containing diverse environments and weather conditions; (3) AVIID, providing aerial infrared-visible imagery for UAV scenarios; (4) LLVIP~\cite{jia2021llvip}, capturing low-light urban scenes; (5) M3FD~\cite{liu2022targetm3fd}, covering multiple weather conditions across various scenes; (6) MSRS~\cite{Ma2022SwinFusion}, containing multi-spectral road scenes; (7) KAIST~\cite{Hwang_2015_CVPRKASIT}, focusing on urban environments; and (8) FLIR~\cite{flir_adas_dataset}, containing thermal and RGB urban street scenes.  

\begin{figure*}[t]
    \centering
    \makebox[\textwidth][c]{\includegraphics[width=\textwidth]{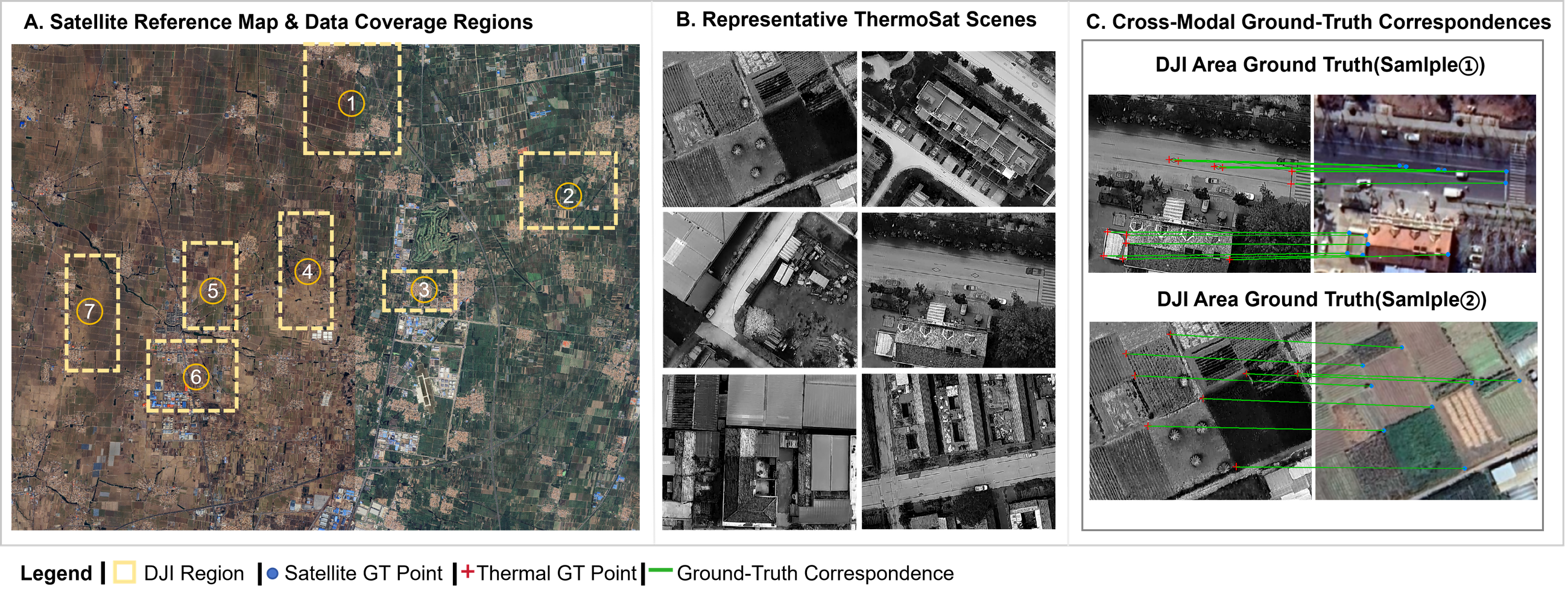}}
    \vspace{-1.0ex}
    \caption{Overview of the proposed ThermoSat dataset. (A) Collection area overview with seven representative thermal-to-satellite localization regions. (B) Representative ThermoSat scenes. (C) Cross-modal ground-truth correspondences, where blue dots denote satellite ground-truth points, red crosses denote thermal ground-truth points, and green lines denote ground-truth correspondences.}
    \vspace{-2.0ex}
    \label{fig:thermosat_overview}
\end{figure*}

Second, for relative pose estimation evaluation, we utilize the METU-VisTIR dataset with relative pose ground truth for building scenes. Third, we contribute our self-collected Aerial Thermal--Satellite Localization Dataset (ThermoSat), containing 832 manually annotated infrared-satellite image pairs captured by a DJI Matrice 4T platform for practical geo-localization evaluation. The thermal images are acquired at a resolution of $1280\times1024$. The selected ThermoSat sequences provide a cumulative trajectory length of approximately 41.56 km, with a total local coverage of approximately 35.86 km$^2$ across the seven surveyed regions. Representative collection areas, thermal scenes, and manually annotated cross-modal correspondences are shown in Fig.~\ref{fig:thermosat_overview}. 

Our framework evaluates 30 algorithms across four key tasks: (1) Homography Estimation; (2) Relative Pose Estimation; (3) Geo-localization; and (4) Hard Geo-localization:

\begin{itemize}

\item \textbf{Homography Estimation}: 
Testing algorithms on the MSCM dataset with random homography transformations to evaluate cross-modal geometric robustness.   

\item \textbf{Relative Pose Estimation}: Using datasets with true camera poses like METU$\_$VisTIR to test the algorithms' performance in 3D scene reconstruction and pose recovery.  

\item \textbf{Geo-localization}: Evaluating practical thermal-to-satellite geo-localization using our self-collected Aerial Thermal--Satellite Localization Dataset (ThermoSat).

\item \textbf{Hard Geo-localization}: Evaluating the same ThermoSat benchmark under a harder protocol constructed from the corresponding ThermoSat sequences by introducing additional flight-induced viewpoint perturbations, including attitude-angle changes and altitude variations, while keeping the underlying thermal-to-satellite localization scenes unchanged.

\end{itemize}

This four-task evaluation framework systematically assesses IR-VIS matching algorithms across a complexity spectrum---from geometric robustness (Homography Estimation) and 3D scene understanding (Relative Pose Estimation) to geo-localization (Geo-localization).

\subsection{Adaptive Preprocessing Front-end}
\begin{figure*}[t]
    \centering
    \includegraphics[width=0.98\textwidth]{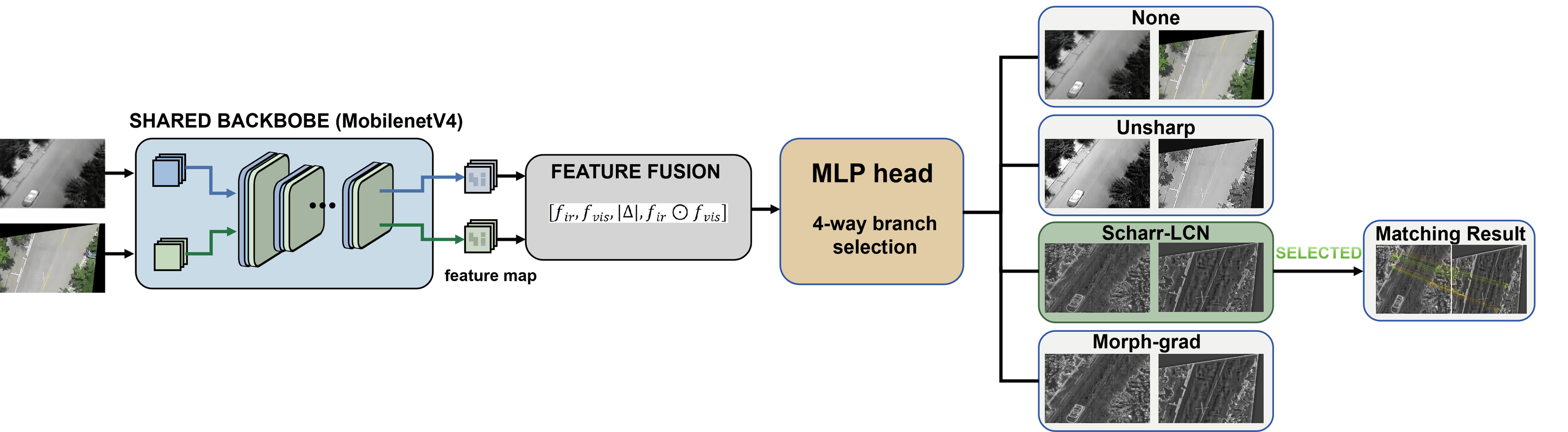}
    \vspace{-1.0ex}
    \caption{Overview of the proposed adaptive preprocessing front-end. A lightweight MobileNetV4-based classifier predicts the most suitable preprocessing branch for each IR-VIS image pair before feature matching.}
    \vspace{-2.0ex}
    \label{fig:adaptive_frontend}
\end{figure*}

Besides benchmarking the matching algorithms themselves, we further introduce an adaptive preprocessing front-end for IR-VIS image pairs. The motivation is that no single preprocessing strategy is consistently optimal for all cross-modal image pairs. Different scenes exhibit different modality gaps in contrast, texture, and structural saliency, while different matching algorithms also respond differently to enhanced inputs. Therefore, a fixed preprocessing branch may improve weak-texture or low-contrast pairs, but it can also reduce performance on image pairs that are already matchable by altering useful intensity patterns or structural details. The overall pipeline is illustrated in Fig.~\ref{fig:adaptive_frontend}.

To address this issue, we cast preprocessing selection as a lightweight image-pair classification problem. Given an infrared-visible image pair, we first resize both images to $224\times224$ and feed them into a classification network built on a MobileNetV4-Conv-Small backbone. The backbone extracts one global feature vector for each image, denoted as $f_{ir}$ and $f_{vis}$. The pair representation is then constructed by concatenating the two original features, their absolute difference $|f_{ir}-f_{vis}|$, and their element-wise product $f_{ir}\odot f_{vis}$. In this way, the fused descriptor jointly encodes single-image appearance cues, cross-modal discrepancies, and shared structural responses. A lightweight classifier then predicts the most suitable preprocessing branch for the current image pair.

The candidate branches are designed to provide complementary properties. The none branch preserves the original image appearance and avoids unnecessary distortion of radiometric and texture information. The unsharp branch strengthens local edges and fine details. The Scharr-gradient with local contrast normalization branch emphasizes gradient structure while suppressing local intensity differences across modalities. The morphological-gradient branch highlights object boundaries and region contours. To train the classifier, we use automatically generated supervision rather than manual branch annotations. Specifically, for each training image pair, we exhaustively evaluate the candidate branches and assign as the target label the branch that produces the largest number of geometric inliers after matching and RANSAC verification~\cite{derpanis2010overview}. The gate network is then optimized as a four-class classifier with the standard cross-entropy loss. The selected branch is finally applied to both modalities before feature extraction and matching, allowing the front-end to adapt to different scene structures and cross-modal appearance gaps. In its current implementation, the gate network contains 7.74M parameters and introduces approximately 21.1\,ms runtime overhead per image pair on a single RTX 4090 GPU at an input resolution of $224\times224$.

\subsection{Evaluation Methodology and Metrics}

\begin{table}[t]
\centering
\caption{Homography estimation results on MSCM. Success rate denotes the ratio of valid homography evaluations over 2000 test pairs, and failed pairs are calculated as zero when computing AUC.}
\renewcommand{\arraystretch}{0.95}
\setlength{\tabcolsep}{3pt}
\resizebox{0.49\textwidth}{!}{
\begin{tabular}{l l c c c c}
\toprule
\rule{0pt}{2.6ex}Category & Method & Success Rate & AUC@5 & AUC@10 & AUC@20 \\
\midrule
\rule{0pt}{2.6ex}Sparse & SuperPoint+MINIMA-LG & 0.987 & \textbf{0.151} & \textbf{0.452} & \textbf{0.689} \\
 & SuperPoint+SuperGlue & 0.971 & \underline{0.114} & \underline{0.358} & \underline{0.559} \\
 & SuperPoint+LightGlue & 0.978 & 0.095 & 0.323 & 0.523 \\
 & ALIKED+LightGlue & 0.926 & 0.072 & 0.246 & 0.414 \\
 & SuperPoint+MNN & \textbf{1.000} & 0.053 & 0.213 & 0.376 \\
 & DeDoDe & \textbf{1.000} & 0.065 & 0.210 & 0.336 \\
 & MIFNet(XFeat) & 0.931 & 0.043 & 0.187 & 0.387 \\
 & XFeat+LightGlue & 0.883 & 0.033 & 0.151 & 0.297 \\
 & D2Net & \textbf{1.000} & 0.030 & 0.151 & 0.326 \\
 & XFeat(sparse) & \textbf{1.000} & 0.029 & 0.133 & 0.277 \\
 & RoRD & \textbf{1.000} & 0.024 & 0.126 & 0.261 \\
 & R2D2 & \underline{0.997} & 0.022 & 0.088 & 0.155 \\
 & SIFT & \textbf{1.000} & 0.018 & 0.066 & 0.107 \\
 & RIFT & 0.988 & 0.004 & 0.024 & 0.068 \\
\midrule
\rule{0pt}{2.6ex}Semi-Dense & XoFTR & \textbf{1.000} & \textbf{0.136} & \textbf{0.414} & \textbf{0.640} \\
 & MINIMA(ELoFTR) & \underline{0.996} & 0.111 & \underline{0.386} & \underline{0.634} \\
 & MINIMA(LoFTR) & \textbf{1.000} & 0.113 & 0.375 & 0.620 \\
 & MINIMA(XoFTR) & 0.935 & \underline{0.126} & 0.366 & 0.573 \\
 & ELoFTR & \textbf{1.000} & 0.100 & 0.336 & 0.551 \\
 & ASpanFormer & \textbf{1.000} & 0.091 & 0.323 & 0.539 \\
 & TopicFM & \textbf{1.000} & 0.089 & 0.299 & 0.509 \\
 & LoFTR & \textbf{1.000} & 0.084 & 0.295 & 0.499 \\
\midrule
\rule{0pt}{2.6ex}Dense & MINIMA(RoMa) & \textbf{1.000} & \textbf{0.198} & \textbf{0.509} & \textbf{0.733} \\
 & RoMa & \textbf{1.000} & \underline{0.151} & \underline{0.438} & \underline{0.672} \\
 & GIM(RoMa) & \textbf{1.000} & 0.114 & 0.353 & 0.580 \\
 & RoMa v2 & \textbf{1.000} & 0.075 & 0.257 & 0.467 \\
 & GIM(DKM) & \textbf{1.000} & 0.076 & 0.245 & 0.405 \\
 & MASt3R & 0.995 & 0.065 & 0.232 & 0.410 \\
 & XFeat(dense) & \underline{0.999} & 0.026 & 0.145 & 0.308 \\
 & DKM & \textbf{1.000} & 0.003 & 0.025 & 0.069 \\
\bottomrule
\end{tabular}
}
\label{tab:auc_homography}  
\end{table}

Our evaluation framework employs multiple quantitative metrics to comprehensively assess the performance of different feature matching algorithms in infrared-visible (IR-VIS) cross-modal scenarios, focusing on four main as follows.  

\subsubsection{Homography Estimation Metrics and Relative Pose Estimation Metrics}  

To evaluate the performance of feature matching methods in geometric tasks, we conduct homography estimation experiments on the MSCM dataset using synthetic geometric transformations. A single random homography transformation is applied to each image pair to provide the corresponding ground-truth homography for evaluation. The transformation parameters are sampled with scale ranging from $0.65$ to $1.35$, rotation ranging from $-25^\circ$ to $25^\circ$, perspective jitter ranging from $-0.23$ to $0.23$, and translation ranging from $-0.17$ to $0.17$ relative to the image width and height, while maintaining at least $60\%$ valid overlap.  
We evaluate homography estimation accuracy using the Area Under the Curve (AUC) with thresholds of 5, 10 and 20 pixels, denoted as AUC@5, AUC@10, and AUC@20, where failed pairs contribute to zero. For relative pose estimation, we also report AUC values at thresholds of 5$^\circ$, 10$^\circ$, and 20$^\circ$. To reduce scene imbalance, we first compute scene-level pose AUC values, then average the scenes within the \textit{cloudy-cloudy} and \textit{cloudy-sunny} splits, and finally combine the two split-level AUCs according to the number of scenes in each split:  

\begin{equation}  
\text{AUC@}\tau = \frac{1}{\tau} \int_{0}^{\tau} \text{Recall}(\epsilon) \, d\epsilon  
\end{equation}  
where $\tau$ represents the threshold value (measured in pixels for homography estimation or in degrees for relative pose estimation), and $\text{Recall}(\epsilon)$ is the proportion of errors below $\epsilon$. For both homography estimation and relative pose estimation, higher AUC values indicate better geometric accuracy and robustness under more stringent evaluation thresholds.

\subsubsection{Geo-localization Metrics}  

For geo-localization and hard geo-localization, we measure localization accuracy in meters using the manually annotated ground-truth correspondences on the geo-referenced image. For each image pair, let $e_i$ denote the localization error in meters, computed as the root-mean-square distance between the projected points and the corresponding ground-truth points. Based on these per-pair errors, we report the median localization error over successfully evaluated pairs, denoted as MedErr$_m$, and the success rates within distance thresholds of 3\,m, 5\,m, and 10\,m over all test pairs, denoted as SR@3m, SR@5m, and SR@10m:  

\begin{equation}
\text{MedErr}_m = \operatorname{median}(\{e_i\}_{i \in \mathcal{S}})
\end{equation}  

\begin{equation}
\text{SR@}\tau = \frac{1}{N}\sum_{i=1}^{N}\mathbf{1}(e_i \le \tau), \quad \tau \in \{3,5,10\}\,\text{m}
\end{equation}  
where $\mathcal{S}$ is the set of successfully evaluated pairs and $N$ is the total number of test pairs. Lower MedErr$_m$ and higher SR@3m/SR@5m/SR@10m indicate better geo-localization performance and stronger robustness in practical deployment.

\begin{table}[t]  
\centering
\caption{Relative pose estimation results on METU-VisTIR. Success Rate denotes the ratio of valid pose evaluations over all 2590 image pairs, and AUC is obtained by combining the \textit{cloudy-cloudy} and \textit{cloudy-sunny} split scores according to the number of scenes in each split.}
\renewcommand{\arraystretch}{0.95}
\setlength{\tabcolsep}{3pt}
\resizebox{0.49\textwidth}{!}{
\begin{tabular}{l l c c c c}
\toprule
\rule{0pt}{2.6ex}Category & Method & Success Rate & AUC@5 & AUC@10 & AUC@20 \\
\midrule
\rule{0pt}{2.6ex}Sparse & SuperPoint+MINIMA-LG & 0.941 & \textbf{0.192} & \textbf{0.373} & \textbf{0.556} \\
 & SuperPoint+SuperGlue & \underline{0.964} & \underline{0.036} & \underline{0.088} & \underline{0.168} \\
 & SuperPoint+LightGlue & 0.941 & 0.016 & 0.043 & 0.090 \\
 & ALIKED+LightGlue & 0.807 & 0.008 & 0.021 & 0.049 \\
 & D2Net & \textbf{1.000} & 0.007 & 0.020 & 0.052 \\
 & RIFT & \textbf{1.000} & 0.006 & 0.018 & 0.048 \\
 & DeDoDe & \textbf{1.000} & 0.004 & 0.010 & 0.029 \\
 & XFeat(sparse) & \textbf{1.000} & 0.003 & 0.010 & 0.033 \\
 & XFeat+LightGlue & 0.410 & 0.001 & 0.005 & 0.017 \\
 & RoRD & \textbf{1.000} & 0.002 & 0.004 & 0.014 \\
 & SuperPoint+MNN & \textbf{1.000} & 0.001 & 0.004 & 0.014 \\
 & R2D2 & \textbf{1.000} & 0.000 & 0.001 & 0.005 \\
 & MIFNet(XFeat) & 0.299 & 0.000 & 0.001 & 0.005 \\
 & SIFT & \textbf{1.000} & 0.000 & 0.000 & 0.002 \\
\midrule
 \rule{0pt}{2.6ex}Semi-Dense & MINIMA(XoFTR) & \underline{0.997} & \textbf{0.197} & \textbf{0.362} & \textbf{0.533} \\
 & XoFTR & \textbf{1.000} & \underline{0.185} & \underline{0.354} & \underline{0.528} \\
 & MINIMA(LoFTR) & \textbf{1.000} & 0.157 & 0.311 & 0.479 \\
 & MINIMA(ELoFTR) & \textbf{1.000} & 0.118 & 0.269 & 0.455 \\
 & ASpanFormer & \textbf{1.000} & 0.041 & 0.098 & 0.194 \\
 & ELoFTR & \textbf{1.000} & 0.038 & 0.096 & 0.201 \\
 & TopicFM & \textbf{1.000} & 0.036 & 0.085 & 0.165 \\
 & LoFTR & \textbf{1.000} & 0.025 & 0.068 & 0.152 \\
\midrule
\rule{0pt}{2.6ex}Dense & MINIMA(RoMa) & \textbf{1.000} & \textbf{0.379} & \textbf{0.612} & \textbf{0.786} \\
 & RoMa & \textbf{1.000} & \underline{0.249} & \underline{0.468} & \underline{0.673} \\
 & RoMa v2 & \textbf{1.000} & 0.225 & 0.436 & 0.635 \\
 & GIM(RoMa) & \textbf{1.000} & 0.165 & 0.374 & 0.594 \\
 & MASt3R & \textbf{1.000} & 0.100 & 0.239 & 0.407 \\
 & DKM & \underline{0.999} & 0.046 & 0.100 & 0.172 \\
 & GIM(DKM) & \textbf{1.000} & 0.000 & 0.030 & 0.129 \\
 & XFeat(dense) & \textbf{1.000} & 0.003 & 0.009 & 0.030 \\
\bottomrule
\end{tabular}
}
\label{tab:auc_comparison}  
\end{table}
%%%%%%%%%%%%%%%%%%%%%%%%%%%%%%%%%%%%%%%%%%%%%%%%%%%%%%%%%%%%%%%%%%%%%%%%%%%%%%%%
\section{Experiments and Results}
We conduct a comprehensive evaluation of 30 state-of-the-art feature matching methods under a unified implementation and metric setting.
For clarity, we abbreviate LightGlue as LG.
\sloppy 
Our benchmark includes 14 sparse methods, namely ALIKED+LG~\cite{aliked,lindenberger2023lightglue}, D2-Net~\cite{Dusmanu2019CVPRd2net}, DeDoDe~\cite{edstedt2024dedode}, MIFNet(XFeat)~\cite{liu2025mifnet,potje2024cvprxfeat}, R2D2~\cite{r2d2}, RIFT~\cite{rift}, RoRD~\cite{parihar2021rord}, SIFT~\cite{sift}, SuperPoint+LG~\cite{superpoint,lindenberger2023lightglue}, SuperPoint+MINIMA-LG~\cite{superpoint,jiang2024minima,lindenberger2023lightglue}, SuperPoint+MNN~\cite{superpoint}, SuperPoint+SuperGlue~\cite{superpoint,sarlin2020superglue}, XFeat(sparse)~\cite{potje2024cvprxfeat}, and XFeat+LG~\cite{potje2024cvprxfeat,lindenberger2023lightglue}; 8 semi-dense methods, namely ASpanFormer~\cite{chen2022aspanformer}, ELoFTR~\cite{wang2024eloftr}, LoFTR~\cite{loftr}, MINIMA(ELoFTR)~\cite{jiang2024minima}, MINIMA(LoFTR)~\cite{jiang2024minima}, MINIMA(XoFTR)~\cite{jiang2024minima}, TopicFM~\cite{giang2023topicfm}, and XoFTR~\cite{tuzcuouglu2024xoftr}; and 8 dense methods, namely DKM~\cite{edstedt2023dkm}, GIM(DKM)~\cite{gim}, GIM(RoMa)~\cite{gim}, MASt3R~\cite{mast3r}, MINIMA(RoMa)~\cite{jiang2024minima}, RoMa~\cite{edstedt2024roma}, RoMa v2~\cite{edstedt2025roma}, and XFeat(dense)~\cite{potje2024cvprxfeat}.  

Note that, MINIMA~\cite{jiang2024minima} and XoFTR~\cite{tuzcuouglu2024xoftr} are specifically designed for IR-VIS matching, while RIFT~\cite{rift} represents a traditional method with cross-modal capabilities. We include MASt3R~\cite{mast3r} instead of DUSt3R due to its superior robustness in challenging scenarios.

\subsection{Experimental Setup}
All experiments are conducted on a workstation equipped with an NVIDIA RTX 4090 GPU. To ensure fair comparison, we standardize the implementation parameters across all methods. Images are resized to maintain a maximum dimension of 640 pixels while preserving their aspect ratio. For feature matching, we set the maximum number of matches to 2048 for all methods. When applying RANSAC for geometric verification, we use a threshold of 3.0 pixels.

For adaptive preprocessing experiments, we employ the MobileNetV4-based front-end described above to select a preprocessing branch for each input pair. The classification network operates on $224\times224$ image pairs and predicts one of four candidate branches, including no preprocessing, unsharp enhancement, Scharr-gradient with local contrast normalization, and morphological-gradient enhancement. The selected preprocessing result is then passed to the downstream matcher for geometric estimation and localization.

\subsection{Results and Discussion}
\label{sec:evaluation}

The evaluation focuses on MSCM for homography estimation, METU-VisTIR for relative pose estimation, and ThermoSat for geo-localization under both base and hard protocols. These results reveal clear differences among sparse, semi-dense, and dense matchers under progressively harder cross-modal geometric tasks.

\subsubsection{Homography Estimation}  
For homography estimation on MSCM, Table~\ref{tab:auc_homography} reports success rate together with AUC@5/10/20. Dense RoMa-family dominate performance on this task. \textbf{MINIMA(RoMa)} achieves the best overall accuracy with AUC@5/10/20 of 0.198/0.509/0.733, followed by \textbf{RoMa} (0.151/0.438/0.672) and \textbf{XoFTR} (0.136/0.414/0.640). Among sparse methods, the  \textbf{SuperPoint+MINIMA-LG} baseline is strongest with AUC@5/10/20 of 0.151/0.452/0.689,  surpassing \textbf{SuperPoint+SuperGlue} and \textbf{SuperPoint+LightGlue}, while handcrafted baselines such as SIFT and RIFT are far behind. \textbf{RoMa v2} attains a 1.000 success rate, while its AUC@10 of 0.257 is still below RoMa, indicating that the gap mainly comes from geometric accuracy rather than failed estimations.

\subsubsection{Relative Pose Estimation}  
For relative pose estimation on METU-VisTIR, Table~\ref{tab:auc_comparison} shows that \textbf{MINIMA(RoMa)} still delivers the best overall accuracy, with AUC@5/10/20 of 0.379/0.612/0.786, followed by \textbf{RoMa} (0.249/0.468/0.673) and \textbf{RoMa v2} (0.225/0.436/0.635). A notable exception appears in the sparse family: \textbf{SuperPoint+MINIMA-LG} reaches 0.192/0.373/0.556, clearly outperforming the other sparse baselines and even surpassing several semi-dense methods on this benchmark. Among the remaining semi-dense matchers, \textbf{MINIMA(XoFTR)} and \textbf{XoFTR} remain the strongest, with AUC@10 of 0.362 and 0.354, respectively.

\subsubsection{Geo-localization}
\begin{table*}[t]
\centering
\caption{Geo-localization results on ThermoSat. Lower MedErr$_m$ and higher SR@3m/SR@5m/SR@10m are better.}
\footnotesize
\renewcommand{\arraystretch}{0.84}
\setlength{\tabcolsep}{2.4pt}
\resizebox{0.90\textwidth}{!}{
\begin{tabular}{l l c c c c c c c c}
\toprule
\rule{0pt}{1.9ex}Category & Method & \multicolumn{4}{c}{Base} & \multicolumn{4}{c}{Hard} \\
\cmidrule(lr){3-6} \cmidrule(lr){7-10}
 &  & MedErr$_m$ & SR@3m & SR@5m & SR@10m & MedErr$_m$ & SR@3m & SR@5m & SR@10m \\
\midrule
\rule{0pt}{1.9ex}Sparse & XFeat(sparse) & 0.746 & \textbf{0.786} & \textbf{0.851} & \textbf{0.921} & 1.417 & \textbf{0.305} & \textbf{0.349} & \textbf{0.398} \\
 & SuperPoint+MINIMA-LG & \textbf{0.452} & \underline{0.686} & \underline{0.719} & 0.758 & \textbf{0.470} & \underline{0.284} & \underline{0.294} & \underline{0.319} \\
 & XFeat+LightGlue & \underline{0.663} & 0.660 & 0.714 & 0.779 & \underline{0.764} & 0.243 & 0.254 & 0.278 \\
 & SuperPoint+SuperGlue & 1.355 & 0.529 & 0.590 & 0.665 & 2.748 & 0.226 & 0.245 & 0.298 \\
 & SuperPoint+LightGlue & 1.429 & 0.513 & 0.573 & 0.655 & 3.404 & 0.208 & 0.234 & 0.282 \\
 & ALIKED+LightGlue & 4.845 & 0.340 & 0.394 & 0.493 & 6.744 & 0.142 & 0.162 & 0.195 \\
 & D2Net & 1.766 & 0.602 & 0.690 & \underline{0.812} & 10.680 & 0.115 & 0.142 & 0.233 \\
 & DeDoDe & 9.312 & 0.242 & 0.314 & 0.519 & 13.050 & 0.085 & 0.107 & 0.180 \\
 & RIFT & 1.965 & 0.554 & 0.626 & 0.749 & 14.674 & 0.064 & 0.091 & 0.145 \\
 & SuperPoint+MNN & 11.276 & 0.188 & 0.264 & 0.448 & 15.420 & 0.061 & 0.089 & 0.154 \\
 & MIFNet(XFeat) & 1.690 & 0.481 & 0.562 & 0.629 & 5.668 & 0.046 & 0.079 & 0.125 \\
 & RoRD & 14.972 & 0.102 & 0.160 & 0.308 & 14.988 & 0.023 & 0.050 & 0.119 \\
 & R2D2 & 16.244 & 0.071 & 0.114 & 0.251 & 18.287 & 0.007 & 0.020 & 0.077 \\
 & SIFT & 18.561 & 0.010 & 0.025 & 0.108 & 17.803 & 0.001 & 0.008 & 0.066 \\
\midrule
\rule{0pt}{1.9ex}Semi-Dense & MINIMA(LoFTR) & \underline{0.456} & \underline{0.847} & \underline{0.905} & \underline{0.950} & \underline{1.201} & \underline{0.291} & \textbf{0.319} & \textbf{0.383} \\
 & MINIMA(ELoFTR) & \textbf{0.371} & \textbf{0.945} & \textbf{0.959} & \textbf{0.977} & \textbf{0.751} & \textbf{0.296} & \underline{0.312} & \underline{0.337} \\
 & ELoFTR & 1.772 & 0.593 & 0.704 & 0.835 & 5.522 & 0.192 & 0.233 & 0.314 \\
 & LoFTR & 2.343 & 0.540 & 0.612 & 0.763 & 6.788 & 0.173 & 0.215 & 0.294 \\
 & MINIMA(XoFTR) & 0.854 & 0.680 & 0.754 & 0.833 & 4.027 & 0.185 & 0.212 & 0.267 \\
 & TopicFM & 2.288 & 0.543 & 0.637 & 0.772 & 6.797 & 0.159 & 0.210 & 0.286 \\
 & XoFTR & 1.528 & 0.582 & 0.671 & 0.793 & 9.174 & 0.153 & 0.188 & 0.252 \\
 & ASpanFormer & 1.091 & 0.694 & 0.787 & 0.874 & 10.535 & 0.129 & 0.159 & 0.228 \\
\midrule
\rule{0pt}{1.9ex}Dense & MINIMA(RoMa) & \textbf{0.343} & \textbf{0.893} & \textbf{0.909} & \underline{0.917} & \textbf{0.333} & \textbf{0.441} & \textbf{0.456} & \textbf{0.472} \\
 & GIM(RoMa) & \underline{0.430} & \underline{0.835} & \underline{0.871} & 0.895 & \underline{0.392} & \underline{0.430} & \underline{0.441} & \underline{0.452} \\
 & RoMa & 0.446 & 0.810 & 0.855 & 0.892 & 0.421 & 0.368 & 0.383 & 0.415 \\
 & XFeat(dense) & 0.760 & 0.797 & 0.869 & \textbf{0.928} & 1.901 & 0.274 & 0.310 & 0.376 \\
 & GIM(DKM) & 1.814 & 0.560 & 0.613 & 0.706 & 1.592 & 0.285 & 0.306 & 0.350 \\
 & DKM & 1.693 & 0.565 & 0.633 & 0.738 & 4.256 & 0.216 & 0.251 & 0.294 \\
 & RoMa v2 & 20.131 & 0.153 & 0.178 & 0.239 & 6.061 & 0.225 & 0.240 & 0.260 \\
 & MASt3R & 18.718 & 0.149 & 0.188 & 0.262 & 20.305 & 0.016 & 0.022 & 0.056 \\
\bottomrule
\end{tabular}
}
\label{tab:thermosat_geo}
\end{table*}

For geo-localization on ThermoSat, Table~\ref{tab:thermosat_geo} summarizes results under both the base and hard protocols. Under the base protocol, \textbf{MINIMA(ELoFTR)} provides the highest threshold-based localization rates, reaching SR@3m/SR@5m/SR@10m of 0.945/0.959/0.977, while \textbf{MINIMA(RoMa)} yields the lowest median error of 0.343 m with SR@5m = 0.909. \textbf{GIM(RoMa)} and \textbf{RoMa} also remain highly competitive with sub-meter median errors and SR@5m above 0.85. Under the hard protocol, all methods degrade markedly, confirming the challenge introduced by the simulated attitude-angle and altitude variations. \textbf{MINIMA(RoMa)} remains the strongest method with MedErr$_m$ = 0.333 m and SR@5m = 0.456, followed by \textbf{GIM(RoMa)} (0.392 m, 0.441) and \textbf{RoMa} (0.421 m, 0.383). Among sparse methods, \textbf{XFeat(sparse)} is the most robust baseline on the hard split, but its localization accuracy still remains below that of the strongest dense methods. \textbf{RoMa v2} again underperforms RoMa on ThermoSat, especially on the base protocol, where its median error rises to 20.131 m.

\subsubsection{Impact of Adaptive Preprocessing Front-end}  
Instead of applying one fixed enhancement pipeline to all samples, we evaluate the proposed MobileNetV4-based adaptive preprocessing front-end using downstream benchmark metrics. For MSCM, we do not train the gate on the full 7083-pair benchmark directly; instead, we first sample a 2000-pair subset from MSCM and then split it into 1600 training pairs and 400 validation pairs. For METU-VisTIR, we use 1382 \textit{cloudy-cloudy} pairs for training and 1208 \textit{cloudy-sunny} pairs for validation. For ThermoSat, we adopt sequence-disjoint splits for both the base and hard protocols. The base protocol uses five sequences for training and two unseen sequences for testing, corresponding to 546 annotated pairs (1092 images) for training and 286 annotated pairs (572 images) for testing. The hard protocol is constructed from the corresponding ThermoSat sequences by introducing additional attitude-angle and altitude perturbations while preserving the same underlying localization scenes, yielding 539 annotated pairs (1078 images) for training and 224 annotated pairs (448 images) for testing.  

\begin{table*}[t]
\centering
\caption{Adaptive preprocessing results on MSCM, METU-VisTIR, and ThermoSat using downstream benchmark metrics. For MSCM and METU-VisTIR, we report AUC@10 on the corresponding validation splits. For ThermoSat, we report SR@10m on the sequence-disjoint hard test split. The ``Gain (\%)'' rows denote the relative percentage change from the baseline to the adaptive result. MSCM results are computed on the 400-pair validation split sampled from the 7083-pair benchmark. METU-VisTIR results are computed on the 1208-pair \textit{cloudy-sunny} validation split. }
\small
\setlength{\tabcolsep}{4pt}
\resizebox{\textwidth}{!}{
\begin{tabular}{llcccccccc}
\toprule
Dataset & Setting & MINIMA(RoMa) & XoFTR & MINIMA(LoFTR) & SuperPoint+LightGlue & TopicFM & LoFTR & DKM & ALIKED+LightGlue \\
\midrule
MSCM & Baseline & 0.526 & 0.421 & 0.391 & 0.354 & 0.325 & 0.311 & 0.308 & 0.292 \\
MSCM & \textbf{Adaptive} & \textbf{0.529} & \textbf{0.426} & \textbf{0.399} & \textbf{0.364} & \textbf{0.356} & \textbf{0.335} & \textbf{0.344} & \textbf{0.328} \\
MSCM & Gain (\%) & +0.6\% & +1.2\% & +2.0\% & +2.8\% & +9.5\% & +7.7\% & +11.7\% & +12.3\% \\
\midrule
METU-VisTIR & Baseline & 0.557 & 0.300 & 0.230 & 0.061 & 0.088 & 0.068 & 0.121 & 0.041 \\
METU-VisTIR & \textbf{Adaptive} & \textbf{0.561} & \textbf{0.312} & \textbf{0.247} & \textbf{0.098} & \textbf{0.135} & \textbf{0.100} & \textbf{0.183} & \textbf{0.088} \\
METU-VisTIR & Gain (\%) & +0.7\% & +4.0\% & +7.4\% & +60.7\% & +53.4\% & +47.1\% & +51.2\% & +114.6\% \\
\midrule
ThermoSat (Hard) & Baseline & 0.478 & 0.281 & 0.393 & 0.286 & 0.299 & 0.286 & 0.304 & 0.174 \\
ThermoSat (Hard) & \textbf{Adaptive} & \textbf{0.485} & \textbf{0.312} & \textbf{0.405} & \textbf{0.301} & \textbf{0.355} & \textbf{0.347} & \textbf{0.333} & \textbf{0.274} \\
ThermoSat (Hard) & Gain (\%) & +1.5\% & +11.0\% & +3.1\% & +5.2\% & +18.7\% & +21.3\% & +9.5\% & +57.5\% \\
\bottomrule
\end{tabular}
}
\label{tab:adaptive_frontend_results}
\end{table*}

Table~\ref{tab:adaptive_frontend_results} reports downstream baseline-versus-adaptive comparisons without mixing in classifier-only diagnostics. On the MSCM validation split, positive gains are observed for all selected matchers, with the improvements appearing for \textbf{DKM}, \textbf{ALIKED+LightGlue}, \textbf{TopicFM}, and \textbf{LoFTR}. On METU-VisTIR, the adaptive front-end also produces consistent gains across all listed matchers, although the improvement for \textbf{MINIMA(RoMa)} is now marginal, while the largest relative gains are still observed for \textbf{ALIKED+LightGlue}, \textbf{SuperPoint+LightGlue}, \textbf{TopicFM}, \textbf{DKM}, and \textbf{LoFTR}. For ThermoSat, we keep only the harder sequence-disjoint protocol in this table. Under this setting, \textbf{MINIMA(RoMa)} remains the strongest selected matcher after adaptation with SR@10m = 0.485, while \textbf{TopicFM}, \textbf{LoFTR}, and \textbf{ALIKED+LightGlue} still show the most notable relative gains of 18.7\%, 21.3\%, and 57.5\%, respectively. These results are consistent with the benchmark observations in the previous subsections: preprocessing preference is strongly coupled with both the scene characteristics and the downstream matcher.

Overall, the adaptive front-end provides a practical way to exploit the fact that the preferred preprocessing branch varies across image pairs and downstream matchers. This observation is consistent with our conjecture: in IR-VIS matching, no single handcrafted preprocessing strategy is universally optimal, while lightweight learned routing can yield measurable gains without changing the matcher architecture itself.

\subsubsection{Efficiency Comparison}
\begin{table}[t]
\centering
\caption{Efficiency comparison at $640\times512$.}
\scriptsize
\renewcommand{\arraystretch}{0.76}
\setlength{\tabcolsep}{2.2pt}
\begin{tabular}{lccc}
\toprule
Method & Params (M) & GFLOPs & Runtime (ms) \\
\midrule
\multicolumn{4}{l}{\textit{Sparse}} \\
SuperPoint+MNN & 1.301 & 111.350 & \textbf{9.29} \\
XFeat(sparse) & 1.545 & \textbf{5.731} & 17.47 \\
SuperPoint+LightGlue & 13.152 & 134.122 & 32.21 \\
XFeat+LightGlue & 1.545 & 26.325 & 37.71 \\
ALIKED+LightGlue & 12.562 & 47.327 & 38.24 \\
SuperPoint+MINIMA-LG & 13.152 & 134.122 & 32.21 \\
SuperPoint+SuperGlue & 13.324 & 129.415 & 54.10 \\
DeDoDe & 28.079 & 2683.537 & 66.66 \\
R2D2 & \textbf{0.484} & 1894.098 & 100.53 \\
SIFT & N/A & N/A & 125.36 \\
D2Net & 7.635 & 724.579 & 188.22 \\
RoRD & 7.635 & 724.579 & 188.57 \\
MIFNet(XFeat) & 15.981 & 52930.926 & 1780.71 \\
RIFT & N/A & N/A & 3667.13 \\
\midrule
\multicolumn{4}{l}{\textit{Semi-Dense}} \\
MINIMA(ELoFTR) & 15.054 & 462.301 & \textbf{34.76} \\
ELoFTR & 15.054 & 462.301 & \textbf{34.76} \\
LoFTR & 11.561 & 763.657 & 42.56 \\
MINIMA(XoFTR) & \textbf{11.092} & 606.850 & 47.47 \\
MINIMA(LoFTR) & 11.561 & 763.657 & 42.56 \\
XoFTR & \textbf{11.092} & 606.850 & 47.47 \\
TopicFM & 11.842 & \textbf{461.943} & 51.36 \\
ASpanFormer & 15.757 & 787.184 & 112.39 \\
\midrule
\multicolumn{4}{l}{\textit{Dense}} \\
XFeat(dense) & \textbf{1.545} & \textbf{21.207} & \textbf{27.54} \\
DKM & 72.261 & 3086.218 & 322.71 \\
RoMa v2 & 425.421 & 13998.896 & 422.70 \\
GIM(DKM) & 70.212 & 5265.993 & 452.97 \\
RoMa & 111.288 & 16347.327 & 614.60 \\
MINIMA(RoMa) & 111.288 & 16347.327 & 614.60 \\
GIM(RoMa) & 111.288 & 16347.327 & 618.61 \\
MASt3R & 688.639 & 14317.709 & 3761.46 \\
\bottomrule
\end{tabular}
\label{tab:efficiency_comparison}
\end{table}

Table~\ref{tab:efficiency_comparison} shows that the sparse family are the fastest methods, with \textbf{SuperPoint+MNN} and \textbf{XFeat(sparse)} providing the lowest runtime. In the semi-dense family, \textbf{MINIMA(ELoFTR)} and \textbf{ELoFTR} offer the best efficiency, while \textbf{XFeat(dense)} is the most lightweight dense method by a clear margin. In contrast, the RoMa-based dense models and \textbf{MASt3R} provide stronger representation capacity at substantially higher computational cost.

%%%%%%%%%%%%%%%%%%%%%%%%%%%%%%%%%%%%%%%%%%%%%%%%%%%%%%%%%%%%%%%%%%%%%%%%%%%%%%%%
\section{Conclusion}
\label{sec:conclusion}
In this paper, we thoroughly test the performance of cross-modal feature algorithms between infrared and visible images. The experiments are conducted using \textbf{30} algorithms across \textbf{10} diverse cross-modal datasets.
We also release ThermoSat, a self-collected infrared-satellite dataset captured by a DJI Matrice 4T platform, and evaluate practical cross-modal visual localization on this dataset.
The evaluation demonstrates MINIMA-RoMa, RoMa, and MINIMA-LG consistently outperform other algorithms across various cross-modal scenarios and tasks. RoMa achieves excellent cross-modal performance by leveraging DINOv2's semantic-rich features that remain consistent to modality differences, while MINIMA series enhances performance through advanced generative models for infrared data synthesis. In addition, the adaptive preprocessing front-end shows that the preferred enhancement branch varies across image pairs and downstream matchers, and that lightweight learned routing can provide consistent downstream gains on MSCM, METU-VisTIR, and the harder ThermoSat protocol without modifying the matcher architecture itself. CM-Bench provides a reliable baseline of cross-modal feature matching and is useful for researchers to design practical cross-modality systems.

\bibliographystyle{IEEEtran}
\bibliography{IEEEabrv,references}

\end{document}